\documentclass[sigconf]{acmart}

\AtBeginDocument{%
  }

\copyrightyear{2025}
\acmYear{2025}
\setcopyright{cc}
\setcctype{by}
\acmConference[SIGIR '25]{Proceedings of the 48th International ACM SIGIR
Conference on Research and Development in Information Retrieval}{July 13--18,2025}{Padua, Italy}
\acmBooktitle{Proceedings of the 48th International ACM SIGIR Conference on Research and Development in Information Retrieval (SIGIR '25), July 13--18, 2025, Padua, Italy}
\acmDOI{10.1145/3726302.3730304}
\acmISBN{979-8-4007-1592-1/2025/07}

\usepackage{graphicx}
\usepackage{makecell} 
\usepackage{xcolor}
\usepackage{booktabs}
\usepackage{multirow}
\usepackage{cleveref} 

\usepackage{balance} 

\begin{document}

\title{Resource for Error Analysis in Text Simplification: New Taxonomy and Test Collection}
\author{Benjamin Vendeville}
\affiliation{%
  \institution{Université de Bretagne Occidentale}
  \institution{Lab-STICC (UMR CNRS 6285)}
  \city{Brest}
  \country{France}
  }
\email{benjamin.vendeville@univ-brest.fr}
\orcid{0009-0003-5298-147X}

\author{Liana Ermakova} 
\affiliation{%
  \institution{Université de Bretagne Occidentale}
  \institution{HCTI, Brest France}
  \city{Brest}
  \country{France}
  }
\email{liana.ermakova@univ-brest.fr}
\orcid{0000-0002-7598-7474}

\author{Pierre De Loor}
\affiliation{%
  \institution{ENIB}
  \institution{Lab-STICC (UMR CNRS 6285)}
  \city{Brest}
  \country{France}
  }
\email{deloor@enib.fr}
\orcid{0000-0002-5415-5505}


\begin{abstract}
  The general public often encounters complex texts but does not have the time or expertise to fully understand them, leading to the spread of misinformation. Automatic Text Simplification (ATS) helps make information more accessible, but its evaluation methods have not kept up with advances in text generation, especially with Large Language Models (LLMs). In particular, recent studies have shown that current ATS metrics do not correlate with the presence of errors. Manual inspections have further revealed a variety of errors, underscoring the need for a more nuanced evaluation framework, which is currently lacking. This resource paper addresses this gap by introducing a test collection for detecting and classifying errors in simplified texts. First, we propose a taxonomy of errors, with a formal focus on information distortion. Next, we introduce a parallel dataset of automatically simplified scientific texts. This dataset has been human-annotated with labels based on our proposed taxonomy. Finally, we analyze the quality of the dataset, and we study the performance of existing models to detect and classify errors from that taxonomy. These contributions give researchers the tools to better evaluate errors in ATS, develop more reliable models, and ultimately improve the quality of automatically simplified texts.
\end{abstract}

\ccsdesc[500]{Information systems~Specialized information retrieval}
\ccsdesc[500]{Information systems~Information extraction}
\ccsdesc[500]{Information systems~Clustering and classification}
\ccsdesc[500]{Information systems~Test collections}

\keywords{Automatic Text Simplification; Error Classification; Hallucinations; Test Collection; Natural Language Processing; Large Language Models}
\maketitle
\section{Introduction}
The internet has made it easier for everyone to share and access information. This has had a big impact both in very general areas, but also in very technical fields such as science. While this democratization has increased availability, it has also highlighted a critical problem: complex language and the lack of background knowledge are challenging obstacles for the general public to understand scientific documents. 
To solve this, the field of Automatic Text Simplification (ATS) aims to develop and investigate methods to facilitate comprehension of complex documents. 
Though not new, the field has grown rapidly with advances in natural language processing and large language models (LLMs)~\cite{openai_GPT4TechnicalReport_2024}. Despite the impressive capabilities of LLMs in text generation, they often introduce various errors~\cite{wu_IndepthEvaluationGPT4_2024}. These errors can include fluency issues (e.g., incorrect syntax, punctuation, or grammar), hallucinations (e.g., generating unsupported or false information)~\cite{goodrich_AssessingFactualAccuracy_2019}, and simplification-specific mistakes such as misrepresenting claims or overgeneralizing concepts~\cite{ermakova_OverviewCLEF2023_2023,ermakova_OverviewCLEF2024_}.

Current efforts in evaluating errors in open-ended generation primarily focus on truthfulness with respect to the source document or world knowledge~\cite{goodrich_AssessingFactualAccuracy_2019,huangSurveyHallucinationLarge2023}. While effective for identifying certain issues, this approach fails to capture the full range of possible errors~\cite{ermakova_OverviewCLEF2023_2023,ermakova_OverviewCLEF2024_}. Error detection 
requires a task-specific approach~\cite{honovich_TRUEReevaluatingFactual_2022}, which has yet to be systematically explored in ATS research. The only related work~\cite{heineman_DancingSuccessFailure_2023} proposes an edit-level evaluation, where some edits (e.g., repetitions, grammar errors, contradictions) are labeled as errors. However, this typology omits certain types of erroneous generations, loosely defines others, and does not align with commonly used terms such as \textit{faithfulness} or \textit{factual} \textit{hallucination}. Besides, the evaluation relies on aligning spans between source and simplified texts, a time-consuming process that becomes even more complex when moving from sentence-level to document-level simplification. These limitations point to the need for a comprehensive taxonomy and annotation framework specifically designed for ATS, which, to our knowledge, has not yet been established.

\textit{Problem.}
The lack of a comprehensive framework for defining, detecting, and evaluating error detection methods in ATS limits progress, new resources are needed.

\begin{figure*}[!t]
  \centering
  \includegraphics[width=0.8\linewidth]{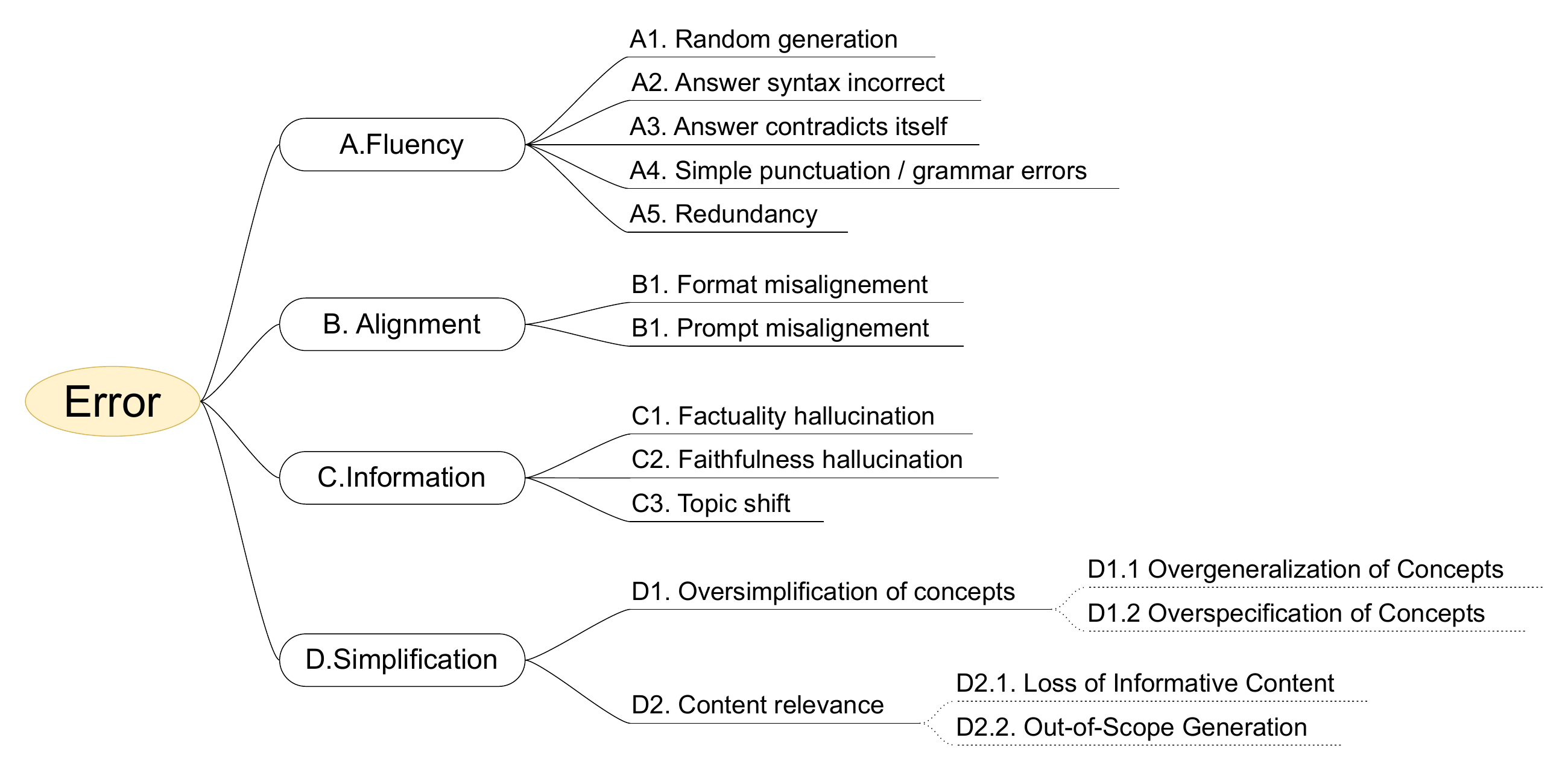}
  \Description[Taxonomy]{Text simplification error typology.}
  \caption{Structure of our typology of errors in text simplification.}
  \label{fig:taxonomy}
\end{figure*}

\textit{Contributions.}
In this resource paper, we make the following contributions to support researchers in building better error detection methods for ATS:
\begin{description}
\item [Taxonomy:]  A new taxonomy of errors in ATS with a focus on information distortion. 
\item [Test Collection:] A test collection based on our taxonomy, accompanied by a detailed annotation scheme.
\item [Showcase:] We showcase error detection methods for identifying and classifying errors using our test collection.
\end{description}
The current state of the art in ATS lacks a formal definition of errors and evaluation tools. To the best of our knowledge, this work provides the first comprehensive taxonomy of ATS errors. In particular, we put an emphasis on formally defining errors related to this task. These resources address key challenges in the field, making it now possible for researchers to more systematically categorize ATS errors and develop more effective tools to detect and avoid them. We will use this test collection to introduce a task on error detection at the SimpleText shared task at CLEF 2025 on automatic text simplification \cite{DBLP:conf/ecir/ErmakovaABVK25}.

The full taxonomy with examples and proper definitions is provided in \hyperref[section:Appendix]{the Appendix}. 
This annotation scheme is designed to be used in a variety of simplification strategies by users with or without expertise in text simplification. The annotated dataset is derived from sentences automatically simplified using various models, which were extracted from system runs submitted to the SimpleText track at CLEF 2024~\cite{ermakova_OverviewCLEF2024_,DBLP:conf/clef/ErmakovaLMK24}.
There is a range of evaluation measures for text simplification against reference simplifications, yet none exhibits very high agreement with human labels of simplification quality~\citep{DBLP:conf/ercimdl/DavariEK24}.  Moreover, earlier analysis revealed a range of information distortion issues in these runs, which were ignored by standard ATS evaluation measures~\citep{DBLP:conf/clef/ErmakovaBMK23,DBLP:conf/clef/ErmakovaLMK24}. Other papers have observed similar issues in the output of generative text simplification models \citep{joseph-etal-2023-multilingual}.

Our resources aim to set a new standard for detecting and evaluating errors in ATS. As IR advances and its applications expand, the demand for tools to manage complex information, especially scientific data, will only increase. We expect ATS research to grow, with our work serving as a foundation for future developments.

In \Cref{section: Developing the Error Taxonomy}, we introduce our taxonomy and the methodology used to develop it, using a formal, fact-based approach to identifying errors in ATS. \Cref{section:Test_Collection} describes our data annotation process based on this taxonomy and presents the results. In \Cref{section: Showcase}, we examine how existing error detection measures perform on our annotated data. We conclude with the implications of these results and a discussion of potential directions for future research.

\section{Developing the Error Taxonomy} \label{section: Developing the Error Taxonomy}

In Text Simplification, the goal is to make text more understandable to a target audience, but this goal may mean different things depending on the audience~\cite{siddharthan_SurveyResearchText_2014}, including different characteristics (syntactic, lexical, …) or different levels (5th grade, 3rd grade, …). We define here errors that appear regardless of the specific simplification goal, as well as errors that are the result of misunderstanding the simplification goal.

\subsection{Structure of the Taxonomy}

The taxonomy is shown in \Cref{fig:taxonomy}. It is a tree composed of the four greater error categories, each with its specific errors.
We define these four greater types of errors, based on where that error comes from, and what made the model make the error:

\begin{itemize}
    \item \textbf{A.\ Fluency.}  Is the answer provided in a correct form that a fluent speaker would speak?
    \item \textbf{B.\ Alignment.} Is the format of the answer correct?
    \item \textbf{C.\ Information.} Is the information provided accurate and relevant to the input?
    \item \textbf{D.\ Simplification.} Does the response focus on simplification?
\end{itemize}

In the following subsections, we discuss each main error category and its subtypes in detail.

\subsection{Fluency Errors}
Errors in this category relate to the capacity of the model to generate human-looking text. The evaluation of the fluency of text is a well-researched area~\cite{laubli_HasMachineTranslation_2018}. In particular, we identify five types of errors that appear in ATS:
\begin{description}
    \item[Random generation] At least part of the answer is just a random string of words or numbers.
    \item[Contradiction] Answer contradicts itself.
    \item[Simple punctuation / grammar errors] Answer has punctuation errors that don't hinder comprehension.
    \item[Redundancy] Repeated sentences, parts of sentences, or groups of sentences that do not need to be repeated.
\end{description}

More details about these errors, including examples, can be found in \hyperref[tax_fluency]{Appendix A.1 Fluency}.

\subsection{Alignment Errors}
Alignment errors occur when the model fails to follow structured prompts or generate responses in the expected format. LLMs learn to interpret tags (e.g., <query>, <answer>), handle one-shot prompts, and maintain proper formatting (e.g., brackets, quotes). Misalignment can cause parsing issues or lead to cascading errors. These errors typically fall into two categories:

\begin{description}
    \item[Format Misalignment] Missing tags or symbols used for formatting, such as JSON symbols (", \}) or starting tags like <answer>.
    \item[Prompt Misalignment] The model generates unnecessary additional content such as another question, a different source text, or an unrelated answer.
\end{description}

More details about these errors, including examples, can be found in \hyperref[tax_alignment]{Appendix A.2 Alignment}.

\subsection{Information Errors} \label{fact_add_sub_section}

\begin{figure}[!t]
    \centering
    \includegraphics[width=1\linewidth]{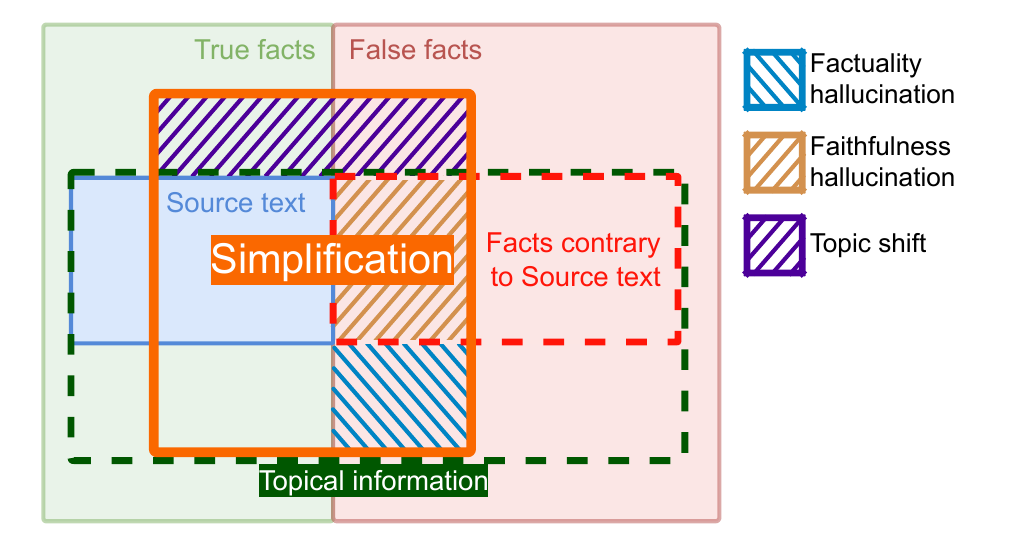}
    \Description[Diagram information errors]{Diagram illustrating the sets and intersections involved in defining information errors.}
    \caption{Diagram illustrating the sets and intersections involved in defining information errors.}
    \label{fig:information_errors_diagram}
\end{figure}

In text simplification, some errors can arise from the model’s treatment of the information in the source text. In particular, we are interested in two questions:
\begin{description}
    \item[Topicality] Is the answer \textit{on topic}?
    \item[Truthfulness] Is the answer \textit{true}?
\end{description}
To answer these questions systematically, we adopt an approach based on sets of facts (defined as tuples of \{subject, relation, target\}) as the unit of information. Let \(F_{src}\) be the set of facts in the source text and \(F_{gen}\) those in the generated text. We have:
\[F_{src} = \{f_s \mid f_s=(subj_s, rel_s, obj_s)\}\]
\[F_{gen} = \{f_g \mid f_g=(subj_g, rel_g, obj_g)\}\]
Where each triplet \((subj, rel, obj)\) consists of a subject, a relation, and an object identified in the source \(S\) or generated text \(G\).
Based on this definition, we represent the sets of facts in \Cref{fig:information_errors_diagram}. \Cref{fig:information_errors_diagram} illustrates the relationship between these sets (\textit{source text} and \textit{simplification}) and the two key questions we explore: \textit{topicality} and \textit{truthfulness}. In the following subsections, we will define and investigate related errors.

\subsubsection{Topicality}

Simplification models have a strong tendency to generate text that contains \textit{off-topic} facts~\cite{ermakova_OverviewCLEF2023_2023}. We define on-topic facts as such:
Let $R_{S}$ denote the set of all subjects, relations, and objects related to the source text topic:
    \[R_{S} = \{ x \mid x \text{ is a subject, relation, or object related to the topic} \}.\]
We define $F_{topic}$ as the set of all facts $(\text{subj}, \text{rel}, \text{obj})$ s.t. $\text{subj}, \text{rel}, \text{obj} \in R_{S}$:
    \[F_{topic} = \{ (\text{subj}, \text{rel}, \text{obj}) \mid \text{subj}, \text{rel}, \text{obj} \in R_{S} \}.\]
From this, we define the following error:
\begin{description}
    \item[Topic Shift] Generated facts that are off-topic. Formally, we define the set of topic shift errors as:
        \[Error_{topic} = F_{gen} \setminus F_{topic}\]
\end{description}

\subsubsection{Truthfulness}

To ensure that the simplified document remains accurate, we examine its truthfulness in relation to two distinct sets of facts:
\begin{align*}
    F_{true} & \quad \text{The set of true facts} \\
    F_{false} & \quad \text{The set of false facts}
\end{align*}

We assume the source document is entirely truthful. A false fact in the simplified text can then arise in two ways:
\begin{description}
    \item[Contradictory to the source] These facts directly contradict information in \(F_{src} \). Formally, we define the set of contradictory facts as:
    \[F_{cont} = \{ f_{cont} \mid \exists f_{src} \in F_{src}, \, f_{src} \text{ contradicts } f_{cont}\}.\]
    \item[Contradictory to general knowledge] These are hallucinated facts that conflict with widely accepted truths.
\end{description}
For simplicity, off-topic facts \(f \notin F_{topic} \) are treated as topicality errors rather than truthfulness errors, even if they are false. We then define two errors:
\begin{description}
    \item[Faithfulness hallucinations] Generated facts that contradict the source document. Formally, we define the set of faithfulness hallucination Errors as:
        \[Error_{faithfulness} = F_{gen} \cap F_{topic} \cap F_{cont}\]
    \item[Factuality hallucinations] Generated facts that contradict general knowledge. Formally, we define the set of factuality hallucination Errors as:
        \[Error_{factuality} = F_{gen} \cap F_{topic} \cap F_{false} \setminus F_{cont}\]
\end{description}
More details about these errors, including examples, can be found in \hyperref[tax:information]{Appendix A.3 Information}.

\subsection{Simplification Errors} \label{Simplification_errors_section}

In text simplification, the goal is to remove unimportant information and add important details (e.g., adding definitions or context) while reformulating the content for simplicity. However, errors can arise when these operations are performed incorrectly. In particular, we are interested in two questions:
\begin{itemize}
    \item Does the answer contain \textit{important} information?
    \item Are the performed \textit{reformulations} correct?
\end{itemize}

As with Information errors, we take a formal approach to studying these questions.
\subsubsection{Importance}

\begin{figure}[!t]
    \centering
    \includegraphics[width=1\linewidth]{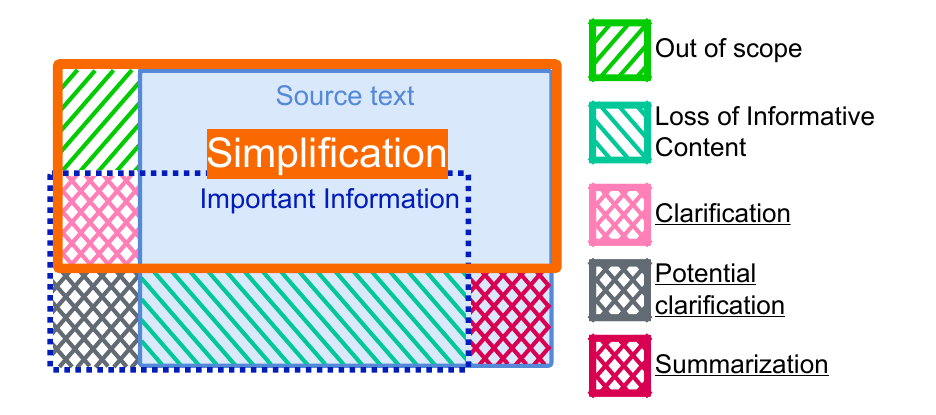}
    \Description[Diagram text simplification errors]{Diagram illustrating the sets and intersections involved in defining simplification errors.}
    \caption{Diagram illustrating the sets and intersections involved in defining simplification errors.}
    \label{fig:importance_errors_diagram}
\end{figure}

Important facts are those that are essential for understanding the central point or purpose of the source document. What we classify as important, however, depends on the goal of the simplification. For instance:
\begin{itemize}
    \item If the goal is to simplify a scientific document for language learners, certain technical details might not be important and could be omitted.
    \item If the goal is to help a non-expert understand the document’s content, these same technical details may become crucial.
\end{itemize}
This highlights that the importance of information varies based on the target audience and purpose. 
As with Information errors, we define \(F_{src}\) as the set of facts in the source text and \(F_{gen}\) as the set of facts in the generated text.
We then consider the set \(F_{imp}\) of facts defined as important by the simplification goal, and make the following assumption:
\begin{itemize}
    \item Every important information is true:
    \( F_{imp} \subset  F_{true} \)
    \item Every important information is on topic: \( F_{imp} \subset  F_{topic} \)
\end{itemize}

\Cref{fig:importance_errors_diagram} illustrates the relationship between these \textit{source text} and \textit{simplification} sets, and importance.
We then define a source text of facts \(F_{src}\) as maximally simple if:
\[F_{src}=F_{imp}\]
We then observe that for every source text that is not maximally simplified, we have one or multiple of the following :
\begin{itemize}
    \item Not every fact included in the source text is necessarily important.  
    \[ \exists f_{src} \in F_{src}, f_{src} \notin F_{imp} \]
    \item There may be important information (e.g. a definition) that is not included in the source text.
    \[ \exists f_{imp} \in F_{imp}, f_{imp} \notin F_{src} \]
\end{itemize}

Therefore, we can identify five different sets:
\begin{itemize}
    \item \textbf{\textit{Out-of-scope generation:}} Facts in the generation that are new and not important
        \[Error_{scope} = F_{gen} \setminus F_{imp} \]
    \item \textbf{\textit{Loss of informative content:}} Source facts that are important and absent from the generation
        \[Error_{Loss} = F_{src} \cap F_{imp} \setminus F_{gen} \]
    \item \textbf{\underline{Summarization}:} Source facts that are not important and removed from the generation
        \[Summ = F_{src} \setminus F_{imp} \setminus F_{gen} \]
    \item \textbf{\underline{Clarification}:} Facts in the generation that are important and new
        \[Clarif = F_{gen} \cap F_{imp} \setminus F_{src} \]
    \item \textbf{\underline{Potential Clarification}:} Facts that are important and not included in the source text or the generation
        \[Clarif_{potential} = F_{imp} \setminus F_{src} \setminus F_{gen} \]
\end{itemize}

Here, we identify two errors (\textit{Out-of-scope generation} and \textit{Loss of informative content}) and three other transformations that do not qualify as errors. The two errors we define here arise from the problem of evaluating what constitutes important information in \textit{the context of the specific simplification goal}. Therefore, we decided to group them under the \textit{Content relevance} subcategory. In the end, these errors will need to be more properly defined based on a proper definition of the needs of each target audience.
The three other transformations (\textit{Summarization}, \textit{Clarification}, and \textit{Potential Clarification}) can be studied as a way to measure the quality of the simplification. 

\subsubsection{Reformulation}
\begin{figure}[!t]
    \centering
    \includegraphics[width=1\linewidth]{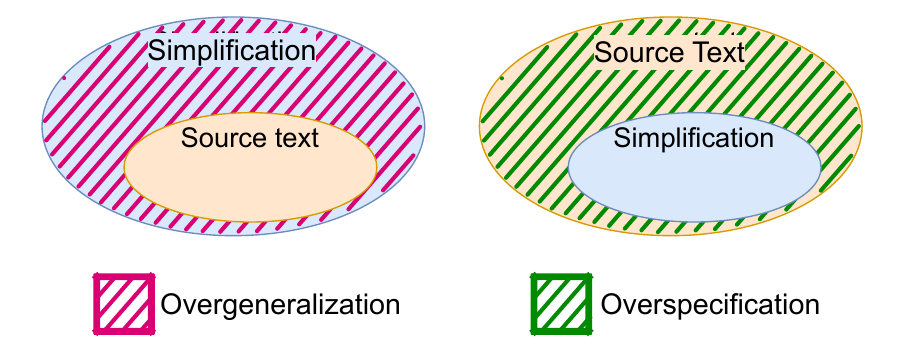}
    \Description[Diagram concept oversimplification errors]{Diagram illustrating the sets and intersections involved in defining concept oversimplification errors.}
    \caption{Diagram illustrating the sets and intersections involved in defining concept oversimplification errors.}
    \label{fig:concepts_diagram}
\end{figure}

Some errors arise not from the \textit{addition} or \textit{subtraction} of facts, but from the \textit{reformulation}. In that context, as shown in \Cref{fig:concepts_diagram}, we may have elements of facts that are replaced by a more general or more specific concept:
\begin{description}
    \item[Generalization] Replacing a concept with a more general one
    \item[Specification] Replacing a concept with a more specific one
\end{description}

More formally, let \(E\) represent the set of elements in a fact: the \textit{subject}, \textit{relation}, or \textit{object}. We define the source fact \(f_{src} = (e_1, e_2, e_{src}) \in F_{src}\) and the generated fact \(f_{gen} = (e_1, e_2, e_{gen}) \in F_{gen}\) such that only one element differs between them. We then have the following substitution errors:
\begin{description} 
    \item[{Overgeneralization}] Replacing a concept with a more general one, which causes the simplified fact to contradict the source. Formally:
        \[ e_{src} \subset e_{gen}, \quad f_{src} \centernot\implies f_{gen} \]
    \item[{Overspecification}] Replacing a concept with a more general one, which causes the simplified fact to contradict the source. Formally:
        \[ e_{src} \supset e_{gen}, \quad f_{src} \centernot\implies f_{gen} \]
\end{description}

A certain level of generalization or specification is often necessary when simplifying text for specific audiences. However, excessive simplification can result in errors, such as overgeneralization or overspecification, where important nuances are lost or altered. To address this, we classify these errors under the broader category of \textit{Oversimplification}. More details about these errors, including examples, can be found in \hyperref[tax:simplification]{Appendix A.4 Simplification}.

\section{Test Collection} \label{section:Test_Collection}

We use this taxonomy to build a test collection of errors in ATS. This collection is composed of a dataset of sentences extracted from scientific abstracts across diverse fields, including medicine, telecoms, and artificial intelligence. The simplifications were gathered during the SimpleText Track at CLEF 2024, where participants then used a variety of methods to generate simplifications, all of them based on LLMs.

We have collected a total of 2,659 individual annotations from ten annotators, as detailed in \Cref{tab:error_distribution}. The annotations reveal a high error rate of 69.16\% in the existing simplifications, with high variability in the occurrence of specific error types. Notably, \textit{Contradiction} appears infrequently (0.86\% occurrence), while \textit{Loss of informative content} is highly prevalent, accounting for 19.56\% of the sentences. These findings show that while text generation models have improved, challenges in fluency remain. Simplification-related errors continue to occur at a high rate, suggesting that further improvements are needed in this area.

\begin{table}[!t]
    \caption{Error Type Distribution: True vs.\ False Counts and Percentage of True Cases}
    \label{tab:error_distribution}
    \setlength\tabcolsep{0pt}
    \begin{tabular*}{\linewidth}{@{\extracolsep{\fill}}lcccc}    
        \toprule
        \textbf{Error Label} & \#\textbf{Total} & \#\textbf{True} & \#\textbf{False} & \%\textbf{True} \\
        \midrule
        No error & 2,659 & 820  & 1,839 & 30.84 \\
\textit{A. Fluency}&&&&\\
        \hspace{2mm}A1. Random generation & 2,659 & 142 & 2,517 & 5.34 \\
        \hspace{2mm}A2. Syntax error & 2,659 & 191 & 2,468 & 7.18 \\
        \hspace{2mm}A3. Contradiction & 2,659 & 23  & 2,636 & 0.86 \\
        \hspace{2mm}\makecell[l]{A4. Punctuation/ \\ grammar error} & 2,659 & 241 & 2,418 & 9.06 \\
        \hspace{2mm}A5. Redundancy & 2,659 & 112 & 2,547 & 4.21 \\
\textit{B. Alignment}&&&&\\
        \hspace{2mm}B1. Format misalignment & 2,659 & 47  & 2,612 & 1.77 \\
        \hspace{2mm}B2. Prompt misalignment & 2,659 & 96 & 2,563 & 3.61 \\
\textit{C. Information}&&&&\\
        \hspace{2mm}C1. Factuality hallucination & 2,659 & 23  & 2,636 & 0.86 \\
        \hspace{2mm}C2. Faithfulness hallucination & 2,659 & 360 & 2,299 & 13.54 \\
        \hspace{2mm}C3. Topic shift & 2,659 & 152 & 2,507 & 5.72 \\
\textit{D. Simplification}&&&&\\
        \hspace{2mm}D1.1. Overgeneralization & 2,659 & 306 & 2,353 & 11.51 \\
        \hspace{2mm}D1.2. Overspecification & 2,659 & 136 & 2,523 & 5.11 \\
        \hspace{2mm}D2.1. Loss of Info.\ Content & 2,659 & 520 & 2,139 & 19.56 \\
        \hspace{2mm}D2.2. Out-of-Scope Gen. & 2,659 & 418 & 2,241 & 15.72 \\
        \bottomrule
    \end{tabular*}
\end{table}

To evaluate inter-annotator agreement, we gathered a separate set of 104 instances, each annotated by five of the annotators. The instances were selected from the subset of data annotated by the lead author to ensure a balanced representation of each error class within the test dataset. This dataset also included eight duplicated instances to measure annotator self-consistency. As shown in \Cref{tab:Consistency_Rate}, the consistency rates varied across annotators, with only two out of five achieving perfect consistency. 
While annotator C achieved low consistency (0.56), we decided to keep their annotations in the dataset to increase the data size. Users who prefer higher quality can choose to exclude their annotations.

\begin{table}[!t]
    \caption{Annotator Consistency Rate}
    \label{tab:Consistency_Rate}
    \setlength\tabcolsep{0pt}
    \begin{tabular*}{\linewidth}{@{\extracolsep{\fill}}lccccc}
    \toprule
        \textbf{Annotator}        & A & B & C & D & E\\
    \midrule
        {Consistency Rate} & 0.78 & 1 & 0.56 & 1 & 0.78 \\
    \bottomrule
    \end{tabular*}
\end{table}

\Cref{tab:error_class_annotators} presents Cohen's Kappa scores for error classes across annotator pairs. Undefined values were set to 1 in cases of complete agreement. Cohen's $\kappa$ indicates the level of inter-annotator agreement, with scores above 0.60 generally considered substantial~\cite{landis_MeasurementObserverAgreement_1977}. Fluency errors show slight to moderate agreement ($\kappa$ = 0.19-0.44) across all pairs. Alignment errors exhibit high variability, with agreement ranging from 0.25 to 1. Information errors shows only slight to fair agreement accross the board ($\kappa$ = 0.06-0.40). Simplification errors scores range from slight to moderate ($\kappa$ = 0.14-49). 

\Cref{tab:error_class_annotators} also displays Fleiss’ Kappa and unanimous agreement. Agreement is highest for Alignment ($\kappa$ = 0.45) and Aggregated A errors ($\kappa$ = 0.38), suggesting clearer definitions or easier detection. In contrast, Information and Simplification errors show lower reliability ($\kappa$ = 0.02–0.26), indicating more subjective or ambiguous judgments.
These results are very diverse and highlight both consistent and challenging areas in error classification.

\begin{table*}[!t]
    \caption{Cohen’s Kappa scores for error classes across all annotator pairs, along with Fleiss’ Kappa scores and percentage of unanimous annotations.}
    \label{tab:error_class_annotators}
    \setlength\tabcolsep{3pt}
    \begin{tabular*}{\linewidth}{@{\extracolsep{\fill}}lcccccccccccc}
        \toprule
        \multirow{2}{*}{\textbf{Error Class}} 
          & \multirow{2}{*}{\textbf{Fleiss' $\kappa$}} 
          & \multirow{2}{*}{\textbf{Unanimous \%}} 
          & \multicolumn{10}{c}{\textbf{Cohen’s Kappa scores for Annotator pair}} \\
        \cmidrule(lr){4-13}
          & & & AB & AC & AD & AE & BC & BD & BE & CD & CE & DE \\
        \midrule
        No error
          & 0.34 & 38.9 & 0.49 & 0.30 & 0.16 & 0.48 & 0.66 & 0.28 & 0.73 & 0.23 & 0.57 & 0.16 \\
        \cmidrule{2-13}
        A. Fluency
          & 0.38 & 67.3 & 0.44 & 0.23 & 0.22 & 0.44 & 0.40 & 0.37 & 0.75 & 0.19 & 0.40 & 0.37 \\
        B. Alignment
          & 0.45 & 76.8 & 0.37 & 0.58 & 1.00 & 0.38 & 0.25 & 0.37 & 0.79 & 0.58 & 0.27 & 0.38 \\
        C. Information
          & 0.02 & 47.3 & 0.06 & 0.22 & 0.14 & 0.07 & 0.17 & 0.40 & 0.38 & 0.04 & 0.27 & 0.28 \\
        D. Simplification
          & 0.26 & 25.2 & 0.26 & 0.24 & 0.14 & 0.40 & 0.49 & 0.19 & 0.45 & 0.20 & 0.45 & 0.12 \\
        \bottomrule
    \end{tabular*}
\end{table*}

This test collection provides a valuable resource for analyzing the challenges of LLM-based text simplification. The high prevalence of errors, particularly in information loss and fluency, underscores the need for further improvements in automatic simplification models. The inter-annotator agreement analysis highlights the importance of annotator training, as expertise significantly impacts the identification of certain error types. By making this test collection available, we aim to facilitate future research in evaluating and enhancing simplification systems, contributing to more reliable and comprehensible scientific text generation.

\section{Showcase} \label{section: Showcase}
Evaluating ATS requires accurate error detection. The question remains, then, whether current error detection models can reliably capture errors as defined by our taxonomy. This showcase will highlight the strengths and limitations of current approaches and guide improvements in ATS evaluation.

\subsection{Methodology}
Our showcase of error detection models focuses on three key questions:
\begin{itemize}
    \item Can they accurately detect the presence of errors in ATS?
    \item Can they effectively identify greater types of errors in ATS?
    \item Can they effectively identify specific types of errors in ATS?
\end{itemize}
Given that hallucination has been one of the most extensively studied errors in recent years~\cite{goodrich_AssessingFactualAccuracy_2019,huangSurveyHallucinationLarge2023} with sometimes imprecise definition of hallucinations, we will use hallucination detection models to evaluate the performance of existing tools. These models are well-suited for identifying factual or faithfulness inconsistencies, making them an ideal starting point for testing the broader framework of error detection in ATS. In addition, we will use  In particular, we will use the following pre-trained models:
\begin{itemize}
    \item Transformer-based models trained on synthetic data
    \begin{itemize}
        \item \textbf{\textit{FactCC}}: Based on BERT, also trained on synthetic summarizations~\cite{kryscinski_EvaluatingFactualConsistency_2020a}.
        \item \textbf{\textit{LENS}}: Based on RoBERTa-large, trained on manually annotated simplifications~\cite{heineman_DancingSuccessFailure_2023}.
    \end{itemize}
    \item Question-Generation \& Question-Answering (QGQA) models
    \begin{itemize}
        \item \textbf{\textit{FEQA}}: Uses two T5 models to assess faithfulness via question-answering~\cite{durmus_FEQAQuestionAnswering_2020a}.
        \item \textbf{\textit{QAGS}}: Another QGQA-based model leveraging two T5 models~\cite{wang_AskingAnsweringQuestions_2020a}.
    \end{itemize}
    \item Fact-based models
    \begin{itemize}
        \item \textbf{\textit{Factacc}}: Uses Named Entity Recognition and a Relation Classifier to verify factual consistency~\cite{goodrich_AssessingFactualAccuracy_2019}.
    \end{itemize}
\end{itemize}
In addition, we will also study the results of BERTScore~\cite{zhangBERTScoreEvaluatingText2020} as it is often used as a benchmark for ATS performance.

To evaluate error detection, since about half of our dataset is error-free, we use AUROC to measure how well models distinguish between correct and erroneous simplifications. For rarer, specific error types, we rely on AUPRC. 
This will help us understand the strengths and limitations of hallucination detection in ATS.

\subsection{Results \& Analysis}

\begin{table}[!t]
    \caption{Greater Error Detection \underline{AUPRC} and Binary Detection "Any Error" \underline{AUROC} for evaluated models.}
    \label{tab:Greater_Errors_auprc}
    \label{tab:binary_auroc}
    \setlength\tabcolsep{0pt}
    \begin{tabular*}{\linewidth}{@{\extracolsep{\fill}}lcccccc}
    \toprule
    \textbf{Metric} & \rotatebox{70}{\textbf{A. Fluency}} & \rotatebox{70}{\textbf{B. Alignment}} & \rotatebox{70}{\textbf{C. Information}} & \rotatebox{70}{\textbf{D. Simplification}}  & \rotatebox{70}{\textbf{No Error}} \\
    \midrule
    \%True      & 21.96         & 5.15          & 19.10         & 43.89          & 30.84 & ~~\\
    \#True      & 584           & 137           & 508           & 1167          & 822\\
    \midrule
    BERTScore   & 0.20          & 0.02          & 0.13          & 0.36          & 0.23 \\
    QAGS        & 0.22          & 0.05          & 0.19          & 0.45          & 0.51 \\
    FEQA        & 0.23          & 0.05          & 0.20          & 0.44          & 0.52 \\
    FactCC      & 0.24          & 0.16          & 0.29          & 0.51          & 0.68 \\
    FactAcc     & 0.24          & 0.05          & 0.19          & 0.42          & 0.44 \\
    LENS        & 0.13          & 0.03          & 0.19          & 0.53          & 0.37\\
    \bottomrule
    \end{tabular*}
\end{table}
\subsubsection{Binary Error Detection AUROC}

\Cref{tab:binary_auroc} summarizes the binary error detection performance as measured by AUROC for the evaluated models. FactCC achieves the highest AUROC of 0.68, indicating a relatively strong ability to distinguish between error and non-error instances. QAGS and FEQA both record moderate performance with AUROC values of 0.51 and 0.52, while FactAcc trails at 0.44. Notably, BERTScore shows a very low AUROC of 0.23, suggesting that its capability for ranking errors in a binary setting is limited.

\subsubsection{Greater Error Type Detection}

\Cref{tab:Greater_Errors_auprc} presents the AUPRC scores for detecting broader error types—fluency, alignment, informativeness, and simplification errors—along with their prevalence. FactCC consistently achieves the highest scores across all categories, with AUPRC of 0.24 for fluency, 0.16 for alignment, 0.29 for information, and 0.51 for simplification. Despite these relative improvements, the overall performance remains low. In particular, detecting Alignment errors is challenging for every model. Surprisingly, even on fluency and alignment errors, which might be 
easier to detect, models exhibit low detection performance.

\subsubsection{Detection for Individual Error Types}

\begin{table*}[!t]
\caption{Caption: Error analysis table presenting various error types and measure scores across multiple metrics. The table reports the AUPRC for BERTScore, QAGS, FEQA, FactCC, FactAcc, and LENS, along with the number of true instances. Errors are grouped into greater categories.}
\label{tab:indiv_errors}
\setlength\tabcolsep{0pt}
\begin{tabular*}{\linewidth}{@{\extracolsep{\fill}}lccccccccc}
\toprule
\makecell[l]{\textbf{Error Type}} & \textbf{\#Total} & \textbf{\#True} & \textbf{\%True} & \textbf{BERTScore ↑} & \textbf{QAGS ↑} & \textbf{FEQA ↑} & \textbf{FactCC ↑} & \textbf{FactAcc ↑} & \textbf{LENS ↑} \\
\midrule

\textit{A Fluency} & & & & & & & & & \\
\hspace{5mm} A1. Random generation & 2,659 & 142 & 5.34 & 0.0275 & 0.0520 & 0.0584 & \textbf{0.1113} & 0.0514 & 0.0288 \\
\hspace{5mm} A2. Syntax error & 2,659 & 191 & 7.18 & 0.0544 & 0.0776 & 0.0831 & \textbf{0.0846} & 0.0716 & 0.0431 \\
\hspace{5mm} A3. Contradiction & 2,659 & 23  & 0.86 & 0.0066 & 0.0118 & 0.0130 & \textbf{0.0208} & 0.0086 & 0.0094 \\
\hspace{5mm} A4. Punctuation/grammar errors & 2,659 & 241 & 9.06 & 0.1371 & 0.0883 & 0.0952 & 0.0710 & \textbf{0.1506} & 0.0542 \\
\hspace{5mm} A5. Redundancy & 2,659 & 112 & 4.21 & 0.0279 & 0.0419 & 0.0453 & \textbf{0.0626} & 0.0404 & 0.0278 \\

\textit{B Alignment} & & & & & & & & & \\
\hspace{5mm} B1. Format misalignement & 2,659 & 47  & 1.77 & 0.0061 & 0.0178 & 0.0176 & \textbf{0.0633} & 0.0170 & 0.0184 \\
\hspace{5mm} B2. Prompt misalignement & 2,659 & 96  & 3.61 & 0.0213 & 0.0349 & 0.0400 & \textbf{0.1200} & 0.0352 & 0.0209 \\

\textit{C Information} & & & & & & & & & \\
\hspace{5mm} C1. Factuality hallucination & 2,659 & 23  & 0.86 & 0.0059 & 0.0099 & \textbf{0.0175} & 0.0105 & 0.0084 & 0.0093 \\
\hspace{5mm} C2. Faithfulness hallucination & 2,659 & 360 & 13.54 & 0.1080 & 0.1421 & 0.1397 & \textbf{0.1704} & 0.1327 & 0.1577 \\
\hspace{5mm} C3. Topic shift & 2,659 & 152 & 5.72 & 0.0306 & 0.0565 & 0.0617 & \textbf{0.1498} & 0.0572 & 0.0485 \\

\textit{D Simplification} & & & & & & & & & \\
\hspace{5mm} D1.1. Overgeneralization & 2,659 & 306 & 11.51 & 0.1011 & 0.1154 & 0.1157 & 0.1210 & 0.1102 & \textbf{0.1871} \\
\hspace{5mm} D1.2. Overspecification & 2,659 & 136 & 5.11 & 0.0404 & 0.0542 & 0.0464 & \textbf{0.0563} & 0.0517 & 0.0540 \\
\hspace{5mm} D2.1. Loss of Informative Content & 2,659 & 520 & 19.56 & 0.1712 & 0.1970 & 0.1848 & 0.2066 & 0.1899 & \textbf{0.2252} \\
\hspace{5mm} D2.2. Out-of-Scope Generation & 2,659 & 418 & 15.72 & 0.1056 & 0.1657 & 0.1803 & \textbf{0.2488} & 0.1517 & 0.2010 \\

\bottomrule
\end{tabular*}
\end{table*}

\Cref{tab:indiv_errors} reports the AUPRC scores for each model across individual error types, grouped into four main categories: Fluency, Alignment, Information, and Simplification. The table also lists the percentage of true instances and the absolute number of true cases for each error type. 
For Fluency errors, FactCC generally achieves the highest AUPRC scores, though the absolute values remain very low. 
A similar trend is observed for Alignment and Information errors, where FactCC outperforms the other models. 
For Simplification errors, the highest scores are shown by LENS. 

Overall, although FactCC consistently shows relatively higher performance, the results always show very poor performance, highlighting the need for improved error detection methods. Moreover, the very low instance counts for certain error types, such as Contradiction, prompt misalignment, and factual hallucination, limit the robustness of the measure.

\section{Conclusion}
We introduced here the first taxonomy of errors in Automatic Text Simplification and built a test collection through the annotation of real-world ATS examples. Our findings show that errors are still prevalent in ATS and that existing methods fail to detect them reliably. While our taxonomy provides a structured approach to error classification, effective annotation requires careful selection and training of annotators. The test collection will be used for the SimpleText shared task at CLEF 2025 and will be published freely after the CLEF 2025 evaluation cycle~\cite{DBLP:conf/ecir/ErmakovaABVK25} in a .csv format on GitHub.\footnote{\url{https://github.com/bVendeville/Salted}}
The repository also provides annotation scheme and the code for analyzing the annotated dataset introduced in this paper. 
By releasing our test collection, taxonomy annotation scheme, and analysis code, we aim to support further research in this area.

Future work could explore automatic evaluation aligned with our taxonomy, improved annotator training, and the use of synthetic data to augment test collections and facilitate the development of ATS models that better mitigate these errors.
%
\begin{acks}
We thank 
master students in translation and technical writing from the University of Brest for their participation in data annotation. 
This research was funded by the French National Research Agency (ANR) under the projects ANR-22-CE23-0019-01 and ANR-19-GURE-0001 (program \emph{Investissements d'avenir} integrated into France~2030).
\end{acks}

\appendix\label{section:Appendix}
\section{Taxonomy}
\subsection{A. Fluency} \label{tax_fluency}
Category focus: is the answer provided in a correct language that a fluent speaker would speak, regardless of the correctness or relevance of the answer?

\underline{\textbf{A1. Random generation}}
\begin{itemize}
    \item \textbf{Definition} At least part of the answer is just a random string of words/numbers
    \item \textbf{Example:}
    \begin{itemize}
        \item \textit{Source:} In the modern era of automation and robotics, autonomous vehicles are currently the focus of academic and industrial research.
        \item \textit{Simplification:} Current academic and industrial research is interested in autonomous vehicles\textbf{.1.2.3.4.5.6.7}
    \end{itemize}
\end{itemize}

\underline{\textbf{A2. Syntax error}}
\begin{itemize}
    \item \textbf{Definition} The syntax is incorrect and doesn't make sense.
    \item \textbf{Example:}
    \begin{itemize}
        \item \textit{Source:} In the modern era of automation and robotics, autonomous vehicles are currently the focus of academic and industrial research.
        \item \textit{Simplification:} In time now of robot and auto, cars that drive self are study much by school and work people.
    \end{itemize}
\end{itemize}

\underline{\textbf{A3. Contradiction}}
\begin{itemize}
    \item \textbf{Definition} Answer contradicts itself.
    \item \textbf{Example:}
    \begin{itemize}
        \item \textit{Source:} In the modern era of automation and robotics, autonomous vehicles are currently the focus of academic and industrial research.
        \item \textit{Simplification:} In today's age of automation and robotics, autonomous vehicles are both widely researched and completely ignored by academics and industry.
    \end{itemize}
\end{itemize}

\underline{\textbf{A4. Simple punctuation or grammar errors}}
\begin{itemize}
    \item \textbf{Definition} Answer has punctuation errors that don't hinder comprehension.
    \item \textbf{Example:}
    \begin{itemize}
        \item \textit{Source:} In the modern era of automation and robotics, autonomous vehicles are currently the focus of academic and industrial research.
        \item \textit{Simplification:} Current academic and industrial research are interested in autonomous vehicles\textbf{…………………}
    \end{itemize}
\end{itemize}

\underline{\textbf{A5. Redundancy}}
\begin{itemize}
    \item \textbf{Definition} Repeated sentences, parts of sentences, or groups of sentences that do not need to be repeated. This is an error regardless of the quality of the sentence.
    \item \textbf{Example:}
    \begin{itemize}
        \item \textit{Source:} In the modern era of automation and robotics, autonomous vehicles are currently the focus of academic and industrial research.
        \item \textit{Simplification:} Current academic and industrial research is interested in autonomous vehicles.\textbf{Current academic and industrial research is interested in autonomous vehicles.}
    \end{itemize}
\end{itemize}

\subsection{B. Alignment} \label{tax_alignment}
Category focus: Does the answer suggest that the model correctly interpreted the prompt, including tags and format?

\underline{\textbf{B1. Format misalignment}}
\begin{itemize}
    \item \textbf{Definition} Some tags or symbols used for formatting are missing. They can include symbols used for JSON parsing (like here with "" and {}) or any "prompt tag" typically <query> <answer> etc…
    \item \textbf{Example:}
    \begin{itemize}
        \item \textit{Source:} In the modern era of automation and robotics, autonomous vehicles are currently the focus of academic and industrial research.
        \item \textit{Simplification:} \{"Current academic and industrial research is interested in autonomous vehicles.\textbf{\}"}
    \end{itemize}
\end{itemize}

\underline{\textbf{B2. Prompt misalignment}}
\begin{itemize}
    \item \textbf{Definition} The model generated one or more of the following:
    \begin{itemize}
        \item unneeded prompt tags (like <query> <answer> etc…) that lead to another question/source etc…
        \item another question (different or not)
        \item another source text (related or not)
        \item another answer (related or not)
    \end{itemize}    
    \item \textbf{Example:}
    \begin{itemize}
        \item \textit{Source:} In the modern era of automation and robotics, autonomous vehicles are currently the focus of academic and industrial research.
        \item \textit{Simplification:} \{"Current academic and industrial research is interested in autonomous vehicles.\textbf{<Query> simplify this: <example>…}
    \end{itemize}
\end{itemize}

\subsection{C. Information}\label{tax:information}
Category focus: does the answer suggest that the model knows and understands everything needed to simplify the input?

\underline{\textbf{C1. Factuality hallucination}}
\begin{itemize}
    \item \textbf{Definition} The simplification contains facts that are contrary to (i.e., can be proven wrong from) "general knowledge" but not directly contrary to the input text.
    \item \textbf{Example:}
    \begin{itemize}
        \item \textit{Source:} In the modern era of automation and robotics, autonomous vehicles are currently the focus of academic and industrial research.
        \item \textit{Simplification:} Current academic and industrial research is interested in autonomous vehicles, which are \textbf{vehicles that can fly}
    \end{itemize}
\end{itemize}

\underline{\textbf{C2. Faithfulness hallucination}}
\begin{itemize}
    \item \textbf{Definition} The simplification contains facts that are contrary to (i.e., can be proven wrong from) the input text.
    \item \textbf{Example:}
    \begin{itemize}
        \item \textit{Source:} In the modern era of automation and robotics, autonomous vehicles are currently the focus of academic and industrial research.
        \item \textit{Simplification:} Current academic and industrial research is \textbf{not at all} interested in autonomous vehicles.
    \end{itemize}
\end{itemize}

\underline{\textbf{C3. Topic shift}}
\begin{itemize}
    \item \textbf{Definition} The generation contains at least some information related to the task (simplification) or the prompt (one-shot encoding) but not to the source document. It can be a text about simplification, or, in the case of one-shot inference, it can be something related to the example given but not to the document that should be simplified.
    \item \textbf{Example 1:}
    \begin{itemize}
        \item \textit{Source:} In the modern era of automation and robotics, autonomous vehicles are currently the focus of academic and industrial research.
        \item \textit{Simplification:} \textbf{Simplification, in the context of language and communication, refers to the process of making text or information easier to understand.}
    \end{itemize}
    \item \textbf{Example 2:}
    \begin{itemize}
        \item \textit{Source:} Simplify the following document: <source>In an attempt to achieve the above-mentioned tasks, we propose an imitation learning based, data-driven solution to UAV  autonomy for navigating through city streets by learning to fly by imitating an expert pilot.<answer>Researchers propose data-driven solutions allowing drones to au\-tonomous\-ly navigate city streets, learning to fly by imitating an expert pilot.<source> In the modern era of automation and robotics, autonomous vehicles are currently the focus of academic and industrial research. <answer>
        \item \textit{Simplification:} \textbf{We propose a data-driven imitation learning method for UAVs to navigate city streets by mimicking an expert pilot.}
    \end{itemize}
\end{itemize}

\subsection{D. Simplification}\label{tax:simplification}
Category focus: Does the answer suggest that the model understands the task of simplification?

\underline{\textbf{D1.1 Overgeneralization of Concepts}}
\begin{itemize}
    \item \textbf{Definition} The simplification removes some precision and generalizes concepts that shouldn't be generalized, making them ambiguous and false. This may include:
    \begin{itemize}
        \item Replacing entities with the greater category of entities
        \item Using vague or ambiguous pronouns in place of clear subjects.
        \item Removing the target of a sentence, implying that a fact applies generally when it only applies in a specific case, to a specific entity
        \item Omitting critical context, such as targets, qualifiers, or conditions.
        \item Generalizing numerical or conditional statements into absolutes.
    \end{itemize}
    \item \textbf{Example 1: Replacing Entities with a Broader Category}
    \begin{itemize}
        \item \textit{Source:} Insects like bees and butterflies are vital for pollination, which is essential for producing many fruits and vegetables.
        \item \textit{Simplification:} Insects are vital for pollination.
    \end{itemize}
    \item \textbf{Example 2: Replacing a Specific Entity with a Pronoun}
    \begin{itemize}
        \item \textit{Source:} The study found that aspirin reduced the risk of heart attack in patients over 50 but had no effect on younger individuals.
        \item \textit{Simplification:} It reduces the risk of heart attack.
    \end{itemize}
    \item \textbf{Example 3: Removing a Target, Leading to Unwarranted Generality}
    \begin{itemize}
        \item \textit{Source:} This vaccine has been shown to be effective in preventing measles in children.
        \item \textit{Simplification:} This vaccine prevents diseases.
    \end{itemize}
\end{itemize}

\underline{\textbf{D1.2. Overspecification of Concepts}}
\begin{itemize}
    \item \textbf{Definition}This error occurs when a broad entity or category in the source text is replaced with a specific example or subcategory during simplification. The source text may intentionally use a general term to avoid unnecessary detail or to maintain flexibility in interpretation. By introducing specificity, the simplified text risks reducing the meaning to an incorrect or unintended entity, misrepresenting the original intent.
    \begin{itemize}
        \item Replacing entities with the greater category of entities
        \item Using vague or ambiguous pronouns in place of clear subjects.
        \item Removing the target of a sentence, implying that a fact applies generally when it only applies in a specific case, to a specific entity
        \item Omitting critical context, such as targets, qualifiers, or conditions.
        \item Generalizing numerical or conditional statements into absolutes.
    \end{itemize}
    \item \textbf{Example:}
    \begin{itemize}
        \item \textit{Source:}  The study examined the impact of climate change on wildlife.
        \item \textit{Simplification:} The study examined the impact of climate change on polar bears.
    \end{itemize}
\end{itemize}

\underline{\textbf{D2.1. Loss of Informative Content}}
\begin{itemize}
    \item \textbf{Definition}Simplifications can omit critical information, making the content uninformative rather than misleading. This omission limits the reader's understanding of the broader context or key points, leaving them unaware of significant elements like parts of a research question, conclusions, or applications. This includes
    \begin{itemize}
        \item A completely empty simplification.
        \item A simplification so general it loses the source's novelty or explanatory value.
        \item Simplifying only one argument when the source has two independent ones.
    \end{itemize}
    \textit{Note}: Deciding what qualifies as "important information" depends on the context. Defining "important information" should consider the audience's needs (e.g., non-native speakers, non-experts, or those with disabilities) and the desired simplicity level (e.g., 3rd grade, 10th grade).
    \item \textbf{Example 1 \textit{Replacing Entities with a Broader Category}:}
    \begin{itemize}
        \item \textit{Source:}  Insects like bees and butterflies are vital for pollination, which is essential for producing many fruits and vegetables.
        \item \textit{Simplification:} Insects are vital for pollination.
    \end{itemize}
    \item \textbf{Example 2 \textit{Replacing a Specific Entity with a Pronoun:}}
    \begin{itemize}
        \item \textit{Source:} The study found that aspirin reduced the risk of heart attack in patients over 50 but had no effect on younger individuals.
        \item \textit{Simplification:}  It reduces the risk of heart attack.
    \end{itemize}
    \item \textbf{Example 3 \textit{Removing a Target, Leading to Unwarranted Generality}:}
    \begin{itemize}
        \item \textit{Source:} This vaccine has been shown to be effective in preventing measles in children.
        \item \textit{Simplification:}  This vaccine prevents diseases.
    \end{itemize}
    \item \textbf{Example 4 \textit{Generalizing Findings Beyond Their Scope}:}
    \begin{itemize}
        \item \textit{Source:} In the controlled study, the intervention improved test scores among high school students in urban areas.
        \item \textit{Simplification:} The intervention improves test scores.
    \end{itemize}
\end{itemize}

\underline{\textbf{D2.2. Out-of-Scope Generation}}
\begin{itemize}
    \item \textbf{Definition} The generation contains information that is unrelated to the task of simplification. The generation may have something to do with the source document to be simplified, but is not about simplifying it. The generation might be:
    \begin{itemize}
        \item An opinion about the source document.
        \item A completion of the source document (more information).
        \item Questions about the source document.
        \item A translation of the source document.
    \end{itemize}
    \item \textbf{Example 1:}
    \begin{itemize}
        \item \textit{Source:} In the modern era of automation and robotics, autonomous vehicles are currently the focus of academic and industrial research.
        \item \textit{Simplification:} Current academic and industrial research is interested in autonomous vehicles. \textbf{In the show KITT with David Hasselhoff the car is an autonomous vehicle and on episode…}
    \end{itemize}
    \item \textbf{Example 2:}
    \begin{itemize}
        \item \textit{Source:} In the modern era of automation and robotics, autonomous vehicles are currently the focus of academic and industrial research.
        \item \textit{Simplification:} Current academic and industrial research is interested in autonomous vehicles. In the show KITT with David Hasselhoff the car is an autonomous vehicle and on episode…
    \end{itemize}
\end{itemize}

\bibliographystyle{ACM-Reference-Format}
\balance
\bibliography{bibliography,anthology}

\balance

\end{document}